\documentclass[conference]{IEEEtran}
\IEEEoverridecommandlockouts
\usepackage{cite}
\usepackage{amsmath,amssymb,amsfonts}
\usepackage{algorithmic}
\usepackage{graphicx}
\usepackage{textcomp}
\usepackage{xcolor}
\usepackage{graphicx}
\def\BibTeX{{\rm B\kern-.05em{\sc i\kern-.025em b}\kern-.08em
    T\kern-.1667em\lower.7ex\hbox{E}\kern-.125emX}}
\begin{document}

\title{Improving Answer Extraction in Context-based Question Answering Systems Using LLMs\\}
% \vspace{0.5cm}
% {\footnotesize \textsuperscript%{*}Note: Sub-titles are not captured in Xplore and
% %should not be used}
% %\thanks{Identify applicable funding agency here. If none, delete this.}
% }

%\and
% \IEEEauthorblockN{Hafez Abdelghaffar Hafez}
% \IEEEauthorblockA{\textit{College of Computer Science} \\
% \textit{MSA University}\\
% Ain Shams, Cairo \\
% hafez.abdelghaffar@msa.edu.eg \\
% Course: NLP (Natural Language Processing) \\
% Instructor: Dr. Ali Hamdy \\}
% \and
%\and
%\and
%\and

\author{\IEEEauthorblockN{ Hafez Abdelghaffar}
\IEEEauthorblockA{\textit{Faculty of Computer Science} \\
\textit{MSA University}\\
Giza, Egypt \\
hafez.abdelghaffar@msa.edu.eg}
\and
\IEEEauthorblockN{ Ahmed Alansary}
\IEEEauthorblockA{\textit{Faculty of Computer Science} \\
\textit{MSA University}\\
Giza, Egypt \\
ahmed.mohamed406@msa.edu.eg}
\and
\IEEEauthorblockN{ Ali Hamdi}
\IEEEauthorblockA{\textit{Faculty of Computer Science} \\
\textit{MSA University}\\
Giza, Egypt \\
ahamdi@msa.edu.eg}}

\IEEEpubid{\makebox[\columnwidth]{ 979-8-3315-8488-7/26/\$31.00 ©2026 IEEE \hfill}
\hspace{\columnsep}\makebox[\columnwidth]{ }}
\maketitle
\IEEEpubidadjcol

\begin{abstract}
Question answering (QA) systems have achieved notable progress with the advent of large language models (LLMs). However, they still face challenges in accurately extracting and generating precise answers from given contexts, particularly when dealing with complex or ambiguous queries. Existing approaches often struggle with contextual understanding, answer consistency, and generalization across diverse domains. In this work, we propose a question answering system based on large language models, where the input consists of a textual context and a corresponding question, and the output is a concise and accurate answer. The motivation behind this research lies in addressing the limitations of current QA systems, particularly their tendency to produce irrelevant or imprecise responses despite having access to the correct context. Our methodology involves fine-tuning a pre-trained LLM on a benchmark QA dataset to improve its contextual comprehension and answer extraction capabilities. Specifically, we utilize the Stanford Question Answering Dataset (SQuAD1.1), which provides high-quality context–question–answer triplets for supervised training and evaluation. Experimental results show that the fine-tuned Roberta-base model achieves the highest performance, attaining a ROUGE-L score of 86.84\%, a BLEU score of 28.24\%, and a BERTScore of 95.38\%. These results indicate strong accuracy and answer relevance, demonstrating the effectiveness of the proposed approach for context-based question answering tasks. Furthermore, the findings confirm that targeted fine-tuning substantially improves the reliability and precision of QA systems.

\end{abstract}


\section{Introduction}

Across a wide range of real-world applications, including search engines, virtual assistants, and customer support platforms, question answering (QA) systems have become increasingly important. As a key area of natural language processing (NLP), these systems allow users to express queries in natural language and receive concise and accurate responses by processing information from both structured and unstructured data sources~\cite{paper15}. Recent progress in large language models (LLMs) and transformer-based architectures has significantly strengthened the capabilities of QA systems. These advances enable such systems to interpret complex queries, capture contextual relationships, and produce more relevant responses. Consequently, QA technologies are increasingly adopted in real-world settings, especially in domains that demand fast and efficient access to extensive textual information.

Although QA systems have achieved substantial progress, their performance continues to be limited by several unresolved challenges. One of the most significant issues is the availability of high-quality training data, since creating large and well-categorized datasets demands considerable time, effort, and financial resources~\cite{paper16}. Real-world questions are often ambiguous or incomplete, making model training more difficult and reducing accuracy. This challenge becomes greater in resource-constrained settings, where linguistic diversity and limited categorized data weaken generalization and increase the need for more robust and adaptable models~\cite{paper17}. Although retrieval-enhanced methods and specialized tools have improved QA system performance, they may limit customization and add complexity to training and deployment~\cite{paper18}. Computational cost, memory usage, and response time also remain key constraints, especially with long contexts or multiple simultaneous queries~\cite{paper19}.

Response quality, consistency, and reliability remain major challenges in QA systems. Noisy or inaccurate inputs may lead to inconsistent answers, while models may rely on stored knowledge instead of the provided context, producing incorrect or misleading responses~\cite{paper21}. In inferential-response tasks, models may identify the correct answer but fail to link it to the most relevant context, reducing response accuracy~\cite{paper22}. LLMs also face difficulties with long contexts, particularly when key information appears in the middle of the text, a challenge known as ``lost in the middle"~\cite{paper23}. Although similar-question generation and knowledge expansion can improve robustness and coverage, they must be carefully designed to balance efficiency and performance~\cite{paper20}. Therefore, these challenges highlight the need for more effective methods to enhance contextual understanding and response accuracy.

This work proposes a question-answering framework that fine-tunes several language models on a benchmark dataset to address these challenges. Using supervised learning and structured context-question-answer triples, the approach aims to strengthen contextual understanding and answer extraction. Multiple models are trained and assessed within the same framework to enable systematic performance comparison and identify the most suitable architectures for QA tasks. Different evaluation metrics are also employed to measure lexical accuracy and semantic similarity. Overall, the framework offers a produce accurate, consistent, and context-relevant answers.

\section{Related Work}

Recent advances in large language models (LLMs) have accelerated the development of question answering (QA) systems, allowing them to process natural language queries and produce concise, contextually appropriate answers \cite{paper7, paper15}. These systems have been applied to both structured and unstructured data sources, supporting diverse applications such as search engines, conversational agents, and educational tools \cite{paper15, paper11}.  Modern frameworks and toolkits also simplify QA system development by combining data collection, preprocessing, fine-tuning, evaluation, and deployment within standardized environments \cite{paper2, paper18}. Despite this progress, developing robust QA systems remains challenging because of costly data annotation, ambiguous natural language queries, and the need for scalable and efficient training methods \cite{paper16, paper7}.

To overcome these limitations, many studies have focused on improving input representations and training methods. Incorporating linguistic features or answer-type information into QA models has been found to enhance performance, especially in resource-limited and linguistically complex settings \cite{paper17, paper25}. Similarly, transforming existing datasets into alternative question formats can lower annotation costs while preserving competitive performance \cite{paper16, paper26}. Despite these advances, LLM-based QA systems continue to face major limitations, such as hallucinations, bias, and substantial computational demands \cite{paper7, paper21}. These limitations become more severe in long-context settings, where models may struggle to relevant information effectively, leading to the ``lost in the middle" problem \cite{paper23, paper27}.

A substantial amount of research has explored the integration of external knowledge sources to enhance QA performance. In particular, knowledge graphs (KGs) offer structured and verifiable information that can improve response accuracy, interpretability, and reliability \cite{paper1, paper12}. Hybrid methods that integrate LLMs with symbolic reasoning systems can also enhance logical consistency and inference quality. By converting textual into formal representations, these approaches support more structured and precise reasoning processes \cite{paper4, paper28}. Moreover, unified frameworks for querying structured data sources have been introduced to handle diverse data types while preserving generalizability and reliability \cite{paper13, paper29}. By addressing the weaknesses of purely generative models, these approaches highlight the value of integrating neural techniques with symbolic reasoning methods.

Another major direction in question-answering research involves retrieval-based methods and retrieval-augmented generation (RAG). By incorporating relevant external documents during inference, these techniques improve answer generation, reduce unsupported guessing, and increase the accuracy of the retrieved information \cite{paper14, paper18}. However, retrieval mechanisms are often vulnerable to bias, noise, and irrelevant contextual information, all of which may adversely affect overall performance \cite{paper14}. To address these limitations, recent studies have introduced advanced retrieval strategies, including reasoning-guided filtering and balanced multi-source retrieval, to enhance both the quality and diversity of the retrieved information \cite{paper14, paper33}. Furthermore, similar-question generation methods and retrieval-based chatbots have been investigated to broaden knowledge bases and enhance system coverage while preserving high levels of reliability and user satisfaction \cite{paper20, paper30}.

A growing research direction explores collaborative, multi-component QA systems, including multi-agent and cumulative frameworks. These approaches decompose the answering process into several stages, such as planning, question understanding, information retrieval, and response generation, which improves the structure and efficiency of the overall reasoning process \cite{paper3, paper10}. Ensemble learning techniques can further enhance performance by combining predictions from multiple models and utilizing their complementary strengths to achieve greater accuracy and reliability \cite{paper8, paper32}. These approaches have shown particular effectiveness in specialized domains such as healthcare, where accurate reasoning and clear interpretability are essential \cite{paper6, paper11}. Nevertheless, adopting these approaches may increase computational demands and architectural complexity, which can restrict their scalability and practical use in real-world applications.

Meanwhile, several studies have aimed to strengthen contextual understanding through improved model architectures and training objectives. Techniques such as block attention and contextual prediction tasks help associate extracted answers with the appropriate context, particularly when a passage contains multiple potential answers \cite{paper22, paper31}. Moreover, previous studies have demonstrated that noisy or conflicting information can reduce the performance of question-answering systems. Many models remain highly sensitive to such factors, which can negatively affect the accuracy, consistency, and overall reliability of their responses \cite{paper21, paper34}. Overcoming these challenges is essential for building dependable question-answering systems that can operate effectively in real-world settings.

Overall, although LLM-based QA systems have advanced considerably, they still face challenges related to reasoning, generalization, contextual understanding, and computational efficiency. These limitations highlight the need for better methods to improve context comprehension and answer extraction. Accordingly, this study evaluates pre-trained language models within a unified framework, aiming to balance accuracy and efficiency in context-based question-answering task.

\begin{table*}[t]
\centering
\renewcommand{\arraystretch}{1.8}
\caption{Sample instances from the Stanford Question Answering Dataset (SQuAD1.1).}
\label{tab:SQuAD1.1_examples}
\begin{tabular}{p{0.60\linewidth} p{0.18\linewidth} p{0.15\linewidth}}
\hline
\textbf{Context} & \textbf{Question} & \textbf{Answer} \\
\hline

Super Bowl 50 was an American football game to determine the champion of the National Football League (NFL) for the 2015 season. The American Football Conference (AFC) champion Denver Broncos defeated the National Football Conference (NFC) champion Carolina Panthers. 
& Who won Super Bowl 50? 
& Denver Broncos \\

\hline

The Amazon rainforest is the largest tropical rainforest in the world, covering much of northwestern Brazil and extending into Colombia, Peru, and other South American countries. It is known for its biodiversity.
& Where is the Amazon rainforest primarily located? 
& Brazil \\

\hline

The Eiffel Tower is a wrought-iron lattice tower on the Champ de Mars in Paris, France. It is named after engineer Gustave Eiffel, whose company designed and built the tower.
& Who designed the Eiffel Tower? 
& Gustave Eiffel \\

\hline

Water boils at 100 degrees Celsius at standard atmospheric pressure. This temperature can vary depending on altitude and pressure conditions.
& At what temperature does water boil under standard conditions? 
& 100 degrees Celsius \\

\hline

The Great Wall of China is a series of fortifications built across the historical northern borders of China to protect against invasions. It is one of the most famous landmarks in the world.
& What was the main purpose of the Great Wall of China? 
& to protect against invasions \\

\hline
\end{tabular}
\end{table*}

\section{Dataset}

The study uses the Stanford Question Answering Dataset (SQuAD1.1) as a benchmark for evaluating machine reading comprehension and inferential reasoning models. The dataset contains paragraphs taken from Wikipedia articles, with questions created by human crowdworkers based on their content. Each question is linked to a specific answer span within the context, enabling the model to demonstrate its ability to understand the passage and extract the correct answer.

SQuAD1.1 includes more than 100,000 question-answer pairs drawn from over 500 Wikipedia articles, covering a broad range of topics. Each sample contains three main elements: the \textit{context}, the \textit{question}, and the corresponding \textit{answer}. Since the answer usually appears as a continuous text span within the context, the dataset is well suited for supervised extractive QA, where the model predicts the start and end positions of the answer.

SQuAD1.1 is generally split into a training set for fine-tuning and a validation set for evaluation and hyperparameter adjustment. Whereas SQuAD1.1 contains only answerable questions, SQuAD2.0 introduces more than 50,000 unanswerable ones, requiring models to decide whether the context provides a valid answer.

SQuAD1.1 is known for its diversity and realistic nature, as its questions are written by human annotators rather than generated automatically. This introduces varied linguistic structures and different reasoning patterns. As a result, the dataset serves as a reliable benchmark for assessing a model's ability to understand context, perform reasoning, and extract answers accurately. Therefore, SQuAD1.1 has become one of the most widely adopted datasets in natural language processing, particularly for developing and evaluating QA systems based on deep learning and large language models.

\section{Methodology}

The proposed question answering (QA) is based on fine-tuning multiple large language models (LLMs). The overall framework follows a supervised learning paradigm where each model learns to map a given textual context and question to a corresponding answer. The system is designed to evaluate and compare the performance of different transformer-based architectures under a unified training setup.

\subsection{Problem Formulation}

Let the dataset be defined as a collection of $N$ training examples:
\begin{equation}
    \mathcal{D} = \{(c_i, q_i, a_i)\}_{i=1}^{N}
\end{equation}

where $c_i$ represents the context paragraph, $q_i$ denotes the associated question, and $a_i$ is the ground truth answer extracted from the context.

\begin{figure}
    \centering
\includegraphics[width=0.7\linewidth]{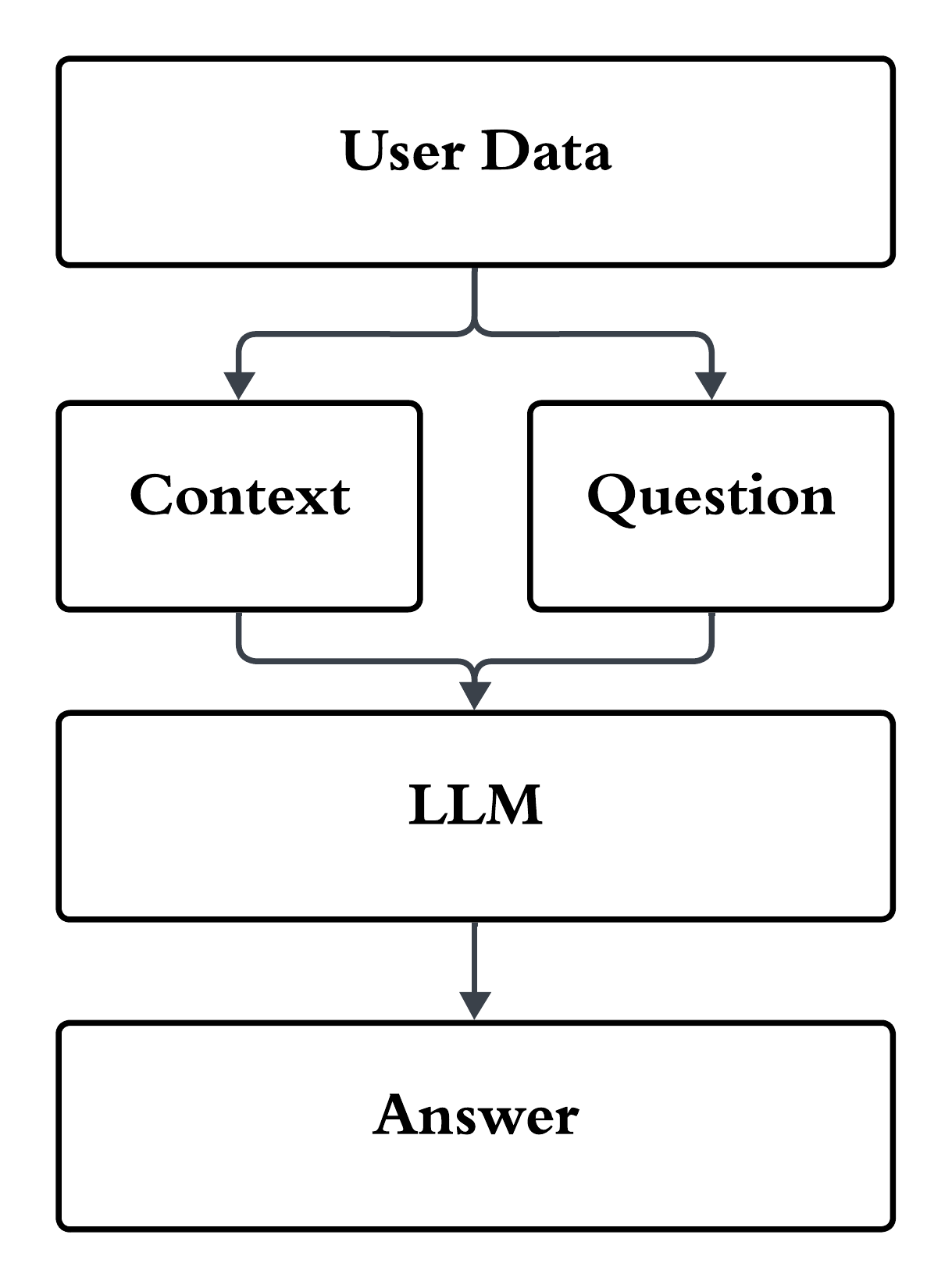}
    \caption{Overview of the proposed context-based question answering system.}
    \label{fig:context}
\end{figure}

As illustrated in Figure \ref{fig:context} the input to the model is constructed by concatenating the context and question:

\begin{equation}
    x_i = [c_i ; q_i]
\end{equation}

The objective of the QA system is to learn a mapping function:

\begin{equation}
    f_\theta : x_i \rightarrow a_i
\end{equation}

where $f_\theta$ is parameterised by $\theta$, representing the learnable weights of the model.

The predicted output is denoted as:

\begin{equation}
    \hat{a}_i = f_\theta(x_i)
\end{equation}

\subsection{Model Architecture}

In this work, transformer-based language models are independently fine-tuned for the QA task.

Each model follows a transformer encoder-based architecture that processes the input sequence $x_i$ and produces contextualised token representations:

\begin{equation}
    H_i = \text{Transformer}_\theta(x_i)
\end{equation}

where $H_i$ denotes the hidden state representations for all input tokens.

\subsection{Answer Representation}

The QA task is formulated as an extractive problem, where the answer is assumed to be a continuous span within the given context. Therefore, the model predicts two values corresponding to the start and end positions of the answer span:

\begin{equation}
    \hat{a}_i = (s_i, e_i)  
\end{equation}

where $s_i$ and $e_i$ denote the predicted start and end indices of the answer within the input sequence.

The probability distributions over start and end positions are computed as:
\begin{equation}
    P_{start}(t|x_i), \quad P_{end}(t|x_i)
\end{equation}

where $t$ represents a token index in the input sequence.

\subsection{Training Objective}

The models are trained using a supervised learning objective that minimises the difference between predicted and ground truth answer spans. The total loss function is defined as:
\begin{equation}
    \mathcal{L} = \mathcal{L}_{start} + \mathcal{L}_{end}
\end{equation}

where:
\begin{equation}
    \mathcal{L}_{start} = -\sum_{i=1}^{N} \log P_{start}(s_i|x_i)
\end{equation}

\begin{equation}
    \mathcal{L}_{end} = -\sum_{i=1}^{N} \log P_{end}(e_i|x_i)
\end{equation}

This formulation encourages the model to assign high probability to the correct start and end positions of the answer span.

\subsection{Fine-Tuning Procedure}

Each of the five LLMs is fine-tuned independently using the same training dataset and identical preprocessing steps. The input format remains consistent across all models, ensuring a fair comparison between architectures.

During training, parameters $\theta$ are optimised using gradient-based optimisation to minimise the loss:
\begin{equation}
    \theta^* = \arg\min_{\theta} \mathcal{L}
\end{equation}

This process allows each model to adapt its pre-trained representations to the QA task while preserving general linguistic knowledge.

\subsection{Evaluation Metrics}

After fine-tuning, each model generates predicted answers $\hat{a}_i$ for the test set. These predictions are compared against the ground truth answers $a_i$ using a set of automatic evaluation metrics that measure lexical overlap and semantic similarity.

The first metric is ROUGE-L, which evaluates the longest common subsequence between the predicted and reference answers, capturing structural similarity:
\begin{equation}
    \text{ROUGE-L}(\hat{a}_i, a_i)
\end{equation}

The second metric is BLEU, which measures n-gram precision between generated and reference answers:
\begin{equation}
    \text{BLEU}(\hat{a}_i, a_i)
\end{equation}

In addition, BERTScore is used to compute semantic similarity using contextual embeddings:
\begin{equation}
    \text{BERTScore}(\hat{a}_i, a_i)
\end{equation}

These metrics collectively provide a comprehensive evaluation of model performance in terms of lexical accuracy, fluency, and semantic alignment.

\subsection{Inference Process}

During inference, the trained model receives a new input pair and generates the predicted answer.

The final answer is extracted from the context using the predicted start and end indices. This ensures that the output remains grounded in the provided passage and maintains factual consistency.

Overall, this methodology enables systematic evaluation of multiple transformer-based models under a unified QA framework, highlighting their effectiveness in contextual understanding, answer extraction, and semantic alignment with ground truth responses.

\section{Results and Discussion}

Experimental results are obtained by evaluating the baseline and fine-tuned models on the QA task. Performance is measured using ROUGE-L, BLEU, and BERTScore, which collectively assess lexical overlap and semantic similarity between the predicted and reference answers.

\begin{table}[t]
\centering
\renewcommand{\arraystretch}{1.5}
\caption{Baseline performance across models.}
\label{tab:baseline_results}
\begin{tabular}{p{0.25\linewidth} p{0.12\linewidth}p{0.15\linewidth}p{0.10\linewidth}p{0.15\linewidth}}
\hline
\textbf{Model} & \textbf{$\#\text{Params}$} & \textbf{ROUGE-L} & \textbf{BLEU} & \textbf{BERTScore} \\
\hline
Bert-tiny & 4.4M & 1.13\% & 0.17\% & 61.41\% \\
TinyLlama-Chat & 1.1B & 8.44\% & 1.59\% & 64.00\% \\
SmolLM2 & 135M & 17.90\% & 4.53\% & 67.74\% \\
Bloom & 1.7B & 9.97\% & 2.14\% & 65.34\% \\
Phi & 1.3B & 22.75\% & 5.63\% & 70.15\% \\
Qwen2.5 & 1.5B & 55.46\% & 18.94\% & 83.59\% \\
Electra-small-discriminator & 14M & 2.69\% & 0.52\% & 62.34\% \\
\textbf{Stablelm-2} & \textbf{1.6B} & \textbf{59.75\%} & \textbf{19.36\%} & \textbf{85.24\%} \\
Distilbert-base-uncased & 66M & 6.98\% & 1.52\% & 62.70\% \\
Bert-base-uncased & 110M & 3.58\% & 0.80\% & 62.77\% \\
Albert-base-v2 & 11.8M & 2.08\% & 0.41\% & 61.56\% \\
Roberta-base & 125M & 3.34\% & 0.58\% & 62.75\% \\
\hline
\end{tabular}
\end{table}

\subsection{Baseline Performance}

Table~\ref{tab:baseline_results} summarises the performance of all models before fine-tuning. Overall, the results indicate generally low to moderate performance across the evaluation metrics, reflecting the limited capability of pre-trained models to perform context-specific question answering without task adaptation. Among the evaluated models, \texttt{Stablelm-2} achieves the highest performance, with a ROUGE-L score of 59.75\%, BLEU of 19.36\%, and BERTScore of 85.24\%. This suggests that larger or more recent architectures may retain stronger generalisation capabilities for QA tasks even without fine-tuning.

It is also observed that BERTScore values are consistently higher than ROUGE-L and BLEU across all models. This indicates that, although the generated answers may not closely match the exact wording of the ground truth, they still retain a certain degree of semantic similarity. However, the overall results confirm that baseline models remain insufficient for accurate QA without fine-tuning.

\subsection{Fine-Tuned Performance}

Table~\ref{tab:finetuned_results} presents the results after fine-tuning the models on the QA dataset. A substantial improvement is observed across all models and evaluation metrics, demonstrating the effectiveness of task-specific training.

The best overall performance is achieved by \texttt{Roberta-base}, which attains the highest scores in all metrics, with ROUGE-L of 86.84\%, BLEU of 28.24\%, and BERTScore of 95.38\%. \texttt{Albert-base} and \texttt{Bert-base} also achieve very high performance, indicating strong capability in accurate answer generation and alignment with the ground truth.

Models such as \texttt{Distilbert-base}, \texttt{Stablelm-2}, and \texttt{Qwen2.5} show strong improvements after fine-tuning, achieving competitive results across all metrics. \texttt{Phi}, \texttt{Bloom}, and \texttt{SmolLM2} also demonstrate substantial gains, highlighting the importance of fine-tuning even for models with moderate baseline performance. Additionally, \texttt{Electra-small} shows a significant improvement, achieving competitive BERTScore. 

Despite these improvements, smaller models such as \texttt{Bert-tiny} continue to lag behind larger counterparts, suggesting that model capacity plays a crucial role in capturing complex contextual relationships required for QA tasks.

\begin{table}[t]
\centering
\renewcommand{\arraystretch}{1.5}
\caption{Fine-tuned performance across models.}
\label{tab:finetuned_results}
\begin{tabular}{p{0.25\linewidth} p{0.12\linewidth}p{0.15\linewidth}p{0.10\linewidth}p{0.15\linewidth}}
\hline
\textbf{Model} & \textbf{$\#\text{Params}$} & \textbf{ROUGE-L} & \textbf{BLEU} & \textbf{BERTScore} \\
\hline
Bert-tiny & 4.4M & 12.00\% & 4.18\% & 65.28\% \\
TinyLlama-Chat & 1.1B & 36.28\% & 11.22\% & 72.26\% \\
SmolLM2 & 135M & 52.63\% & 20.58\% & 82.72\% \\
Bloom & 1.7B & 53.17\% & 18.98\% & 83.38\% \\
Phi & 1.3B & 58.49\% & 20.20\% & 84.20\% \\
Qwen2.5 & 1.5B & 64.91\% & 23.21\% & 85.87\% \\
Electra-small-discriminator & 14M & 61.52\% & 20.64\% & 85.93\% \\
Stablelm-2 & 1.6B & 65.41\% & 22.46\% & 86.45\% \\
Distilbert-base-uncased & 66M & 69.34\% & 22.51\% & 89.02\% \\
Bert-base-uncased & 110M & 78.00\% & 26.24\% & 92.38\% \\
Albert-base-v2 & 11.8M & 82.27\% & 26.56\% & 93.85\% \\
\textbf{Roberta-base} & \textbf{125M} & \textbf{86.84\%} & \textbf{28.24\%} & \textbf{95.38\%} \\
\hline
\end{tabular}
\end{table}

\subsection{Comparative Analysis}

A comparison between Tables~\ref{tab:baseline_results} and~\ref{tab:finetuned_results} clearly highlights the impact of fine-tuning. All models exhibit substantial increases in ROUGE-L, BLEU, and BERTScore, confirming that supervised training enables better alignment between predicted and reference answers.

In particular, ROUGE-L and BLEU scores show the most significant improvements, indicating enhanced lexical overlap and more precise answer generation. BERTScore also improves consistently across all models, suggesting that fine-tuned models produce answers that are not only lexically accurate but also semantically meaningful.

Furthermore, the relative ranking of models changes notably after fine-tuning. While \texttt{Stablelm-2} and \texttt{Qwen2.5} perform best in the baseline setting, they are surpassed by encoder-based models such as \texttt{Roberta-base}, \texttt{Albert-base}, and \texttt{Bert-base} after fine-tuning. This indicates that pre-training alone is not a reliable indicator of QA performance, and that adaptability to the task plays a critical role.

Overall, the results demonstrate that fine-tuning is essential for achieving high-performance QA systems. The findings also suggest that, although smaller models can benefit from fine-tuning, larger and more expressive architectures tend to deliver superior results in terms of both accuracy and semantic quality.

\section{Conclusion}

This work presented a question answering (QA) framework based on fine-tuning multiple large language models (LLMs) on a benchmark dataset. The study aimed to improve contextual understanding and answer extraction by adapting pre-trained models to the QA task using supervised learning. A unified training and evaluation setup was employed to ensure a fair comparison across different model architectures.

The experimental results demonstrated that fine-tuning significantly enhances model performance across all evaluation metrics, including ROUGE-L, BLEU, and BERTScore. All models showed substantial improvements compared to their baseline counterparts, confirming the importance of task-specific training. In particular, models such as \texttt{Albert-base} and \texttt{Bert-base} achieved the highest performance, indicating their strong capability in capturing contextual relationships and generating accurate answers.

The comparative analysis further revealed that model capacity and architecture play a critical role in QA performance. While lightweight models benefit from fine-tuning, they generally lag behind larger models in handling complex queries and producing precise responses. Additionally, the results highlighted that relying solely on pre-trained knowledge is insufficient for high-quality QA, and that fine-tuning is essential for achieving reliable and consistent outputs.

Overall, the findings confirm that leveraging LLMs with targeted fine-tuning provides an effective solution for context-based question answering tasks. Future work may explore advanced techniques such as retrieval-augmented generation, multi-model fusion, or domain-specific adaptation to further enhance performance and robustness across diverse QA scenarios.

\bibliographystyle{IEEEtran}
\bibliography{ref}

\end{document}